%% file: accv2022submission.tex
\documentclass[runningheads]{llncs}
\usepackage{graphicx}
\usepackage{amsmath,amssymb} 
\usepackage{color}

\usepackage{url}
\usepackage[labelformat=simple]{subcaption}

\newcommand{\softmax}{\operatorname{softmax}}
\newcommand{\head}{\operatorname{head}}

\newcommand{\R}{\mathbb{R}}

\begin{document}
\pagestyle{headings}
\mainmatter

\def\ACCV22SubNumber{2}  

\title{Temporal Cross-attention for Action Recognition} 
\titlerunning{Temporal Cross-attention for Action Recognition}
\authorrunning{Temporal Cross-attention for Action Recognition}


\author{Ryota Hashiguchi \and Toru Tamaki\orcidID{0000-0001-9712-7777}}
\institute{Nagoya Institute of Technology, Nagoya, Japan\\
\email{r.hashiguchi.651@nitech.jp, tamaki.toru@nitech.ac.jp}
}

\maketitle

\begin{abstract}
\input{abstract}
\end{abstract}

\input{main_text.tex}

\subsubsection*{Acknowledgements.}
This work was supported in part by JSPS KAKENHI Grant Number JP22K12090.

\bibliographystyle{splncs}
\bibliography{mybib}

\end{document}

%% file: abstract.tex

Feature shifts have been shown to be useful for action recognition with CNN-based models since Temporal Shift Module (TSM) was proposed.
It is based on frame-wise feature extraction with late fusion, and layer features are shifted along the time direction for the temporal interaction.
TokenShift, a recent model based on Vision Transformer (ViT), also uses the temporal feature shift mechanism, which, however, does not fully exploit the structure of Multi-head Self-Attention (MSA) in ViT.
In this paper, we propose \emph{Multi-head Self/Cross-Attention} (MSCA),
which fully utilizes the attention structure.
TokenShift is based on a frame-wise ViT with features temporally shifted with successive frames (at time $t+1$ and $t-1$).
In contrast, the proposed MSCA replaces MSA in the frame-wise ViT, and some MSA heads attend to successive frames instead of the current frame.
The computation cost is the same as the frame-wise ViT and TokenShift as it simply changes the target to which the attention is taken.
There is a choice about which of key, query, and value are taken from the successive frames,
then we experimentally compared these variants with Kinetics400.
We also investigate other variants in which the proposed MSCA is used along the patch dimension of ViT, instead of the head dimension.
Experimental results show that a variant, MSCA-KV, shows the best performance and is better than TokenShift by 0.1\% and then ViT by 1.2\%.

%% file: main_text.tex
\section{Introduction}

Recognizing the actions of people in videos is an important topic in computer vision.
After the emergence of Vision Transformer (ViT) \cite{ViT}
which has been shown to be effective for various image recognition tasks \cite{CLIP,DALL-E},
research extending ViT to video has become active \cite{ViViT,VidTr,Video-Transformer-Network,TimeSformer,STAM}.

In recognition of video, unlike images,
it is necessary to model the temporal information across frames of a given video clip.
While most early works of action recognition were frame-wise CNN models followed by temporal aggregation
\cite{Karpathy_2014_CVPR,Donahue_2015_CVPR},
3D CNN models \cite{C3D,I3D,R3D} were shown to be effective
and well generalized when they were
trained on large-scale datasets \cite{Kinetics} and 
transferred to smaller datasets \cite{UCF101}.

However, the computational cost of 3D convolution is usually high.
To mitigate the issue of the trade-off between the ability of temporal modeling and the high computational cost,
Temporal Shift Module (TSM) \cite{Lin_2019_ICCV} was proposed by extending spatial feature shifting \cite{CNN-Transformer-CVPR2019,Wu_2018_CVPR} to the temporal dimension.
TSM uses late fusion by applying 2D ResNet \cite{ResNet2015} to each frame,
while it exploits the temporal interaction between adjacent frames by shifting the features in each layer of ResNet.
The feature shift operation is computationally inexpensive,
nevertheless it has been shown to perform as well as 3D CNN.
Therefore, many related CNN models \cite{Gated-Shift-Network,LGTSM,RubiksNet} have been proposed.
TokenShift \cite{Token-Shift-Transformer} is a TSM-like 2D ViT-based model that shifts a fraction of ViT features,
showing that shifting only the class tokens is enough to achieve a good performance.

However, TokenShift doesn't fully exploit the advantage of the architecture of the attention mechanism of ViT,
instead it simply shifts features like as in TSM.
We propose a method that seamlessly utilizes the ViT structure to interact with features across adjacent frames. 
The original TSM or TokenShift uses a 2D model applied to each frame,
while the output features of layers of the 2D model are then shifted from the current time $t$ to one time step forward $t+1$ and backward $t-1$.
In the encoder architecture of ViT,
Multi-head Self-Attention (MSA) computes attentions between patches at a given frame $t$.
The proposed method utilizes and modifies it for the temporal interaction;
some patches at frame $t$ attends to 
not only patches at the current frame $t$
but also patches at the neighbor frames $t+1$ and $t-1$.
We call this mechanisms \emph{Multi-head Self/Cross-Attention} (MSCA).
This allows the temporal interaction to be computed within the Transformer block without additional shift modules.
Furthermore, the computational cost remains the same because some parts of self-attention computation is simply replaced with the temporal cross-attention.
The contributions of this paper are as follows:
\begin{itemize}
\item We propose a new action recognition model with the proposed MSCA, which seamlessly combines the concepts of ViT and TSM.
This avoids computing spatio-temporal attention directly in the 3D video volume.
\item MSCA uses cross-attention to perform temporal interactions inside the Transformer block,
unlike TokenShift that uses additional shifting modules.
This allows for temporal interaction without increasing computational complexity nor changing model architecture.
\item Experiments on Kinetics400 show that the proposed method improves performance compared to ViT and TokenShift.
\end{itemize}

\section{Related Work}

\subsection{2D/3D CNN, and TSM}

Early approaches of deep learning to action recognition are 2D-based methods \cite{Karpathy_2014_CVPR,Donahue_2015_CVPR}
that apply 2D CNN to videos frame by frame,
or Two-Stream methods \cite{two-stream}
that use RGB frames and optical flow stacks.
However, it is difficult to model long-term temporal information beyond several time steps,
and this is where 3D-based methods \cite{C3D,I3D,R3D} came in.
These models simultaneously consider spatial and temporal information,
but their computational cost is relatively high compared to 2D-based methods.
Therefore, 
mixed models of 2D and 3D convolution \cite{P3D,S3D,R(2+1)D} have been proposed by separating convolution in the spatial and temporal dimensions.


TSM \cite{Lin_2019_ICCV} is an attempt to model temporal information for action recognition without increasing parameters and computational complexity while maintaining the architecture of 2D CNN-based methods.
TSM shifts intermediate features between neighbor frames, allowing 2D CNNs of each frame to interact with other frames in the temporal direction.
Many CNN-based variants have followed,
such as Gated-Shift Network \cite{Gated-Shift-Network} that learns temporal features using only 2D CNN with the gate shift module, and RubiksNet \cite{RubiksNet} which uses three learnable shifts (spatial, vertical, and temporal) for end-to-end learning.

\subsection{ViT-based video models}

ViT \cite{ViT} performs very well in image recognition tasks, and its application to action recognition tasks has been actively studied \cite{ViViT,VidTr,Video-Transformer-Network,TimeSformer,STAM,tokenlearner,bulat2021spacetime}.
It is known that the computational cost of self-attention in Transformer is $O(N^2)$ where $N$ is the number of tokens,
or the number of patches $N$ for ViT.
For videos, the number of frames $T$ also contributes to the computational cost of spatio-temporal self-attention.
A simple extension of ViT with full spatio-temporal attention gives a computational cost of $O(N^2 T^2)$,
since the number of tokens increases in proportion to the temporal extent.
To alleviate this computational issue,
TimeSformer \cite{TimeSformer} reduces the computational complexity to $O(T^2N+TN^2)$ by applying temporal and spatial attention separately,
TokenLeaner \cite{tokenlearner} to $O(S^2T^2)$ by selecting $S (<N)$ patches that are important for the representation, and
Space-time Mixing Attention \cite{bulat2021spacetime} to $O(T N^2)$ by attending tokens in neighbor frames only.

TokenShift \cite{Token-Shift-Transformer} is a TSM-like model that shifts a fraction of ViT features in each Transformer block.
It computes the attention only in the current frame, hence the complexity is $O(T N^2)$ with fewer FLOPs than \cite{bulat2021spacetime}.
TokenShift shifts the class token only based on experimental results,
however it doesn't fully exploit the attention mechanism.
In contrast, the proposed method makes use of the attention mechanism for feature shift; we replace some of the self-attention modules with the proposed cross-attention modules that shift key, query, and value to neighbor frames, instead of naively shifting features.

\section{Method}

\begin{figure}[t]

    \centering

    \begin{minipage}{.15\linewidth}
    \centering
    \includegraphics[width=\linewidth]{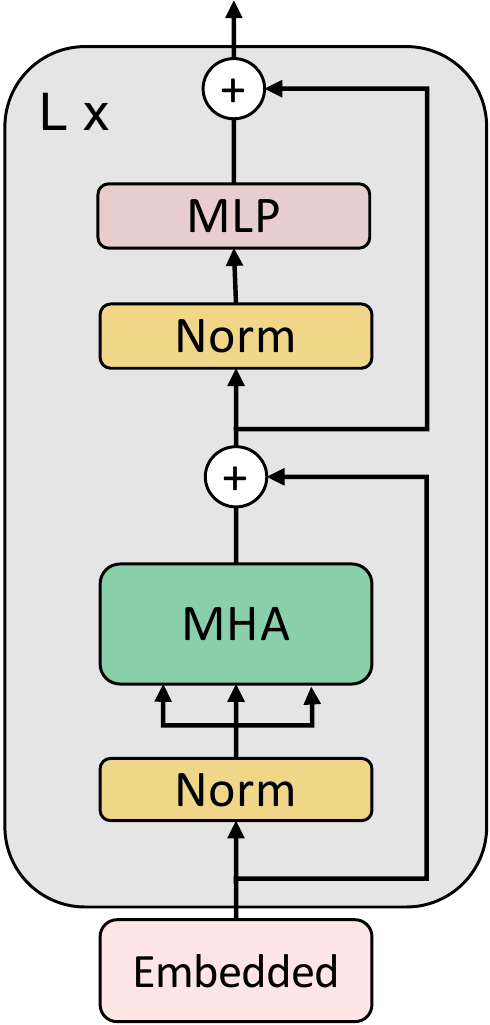}\\
    \subcaption{}
    \label{fig:ViT_block}
    \end{minipage}
    \hfil
    \begin{minipage}{.15\linewidth}
    \centering
    \includegraphics[width=\linewidth]{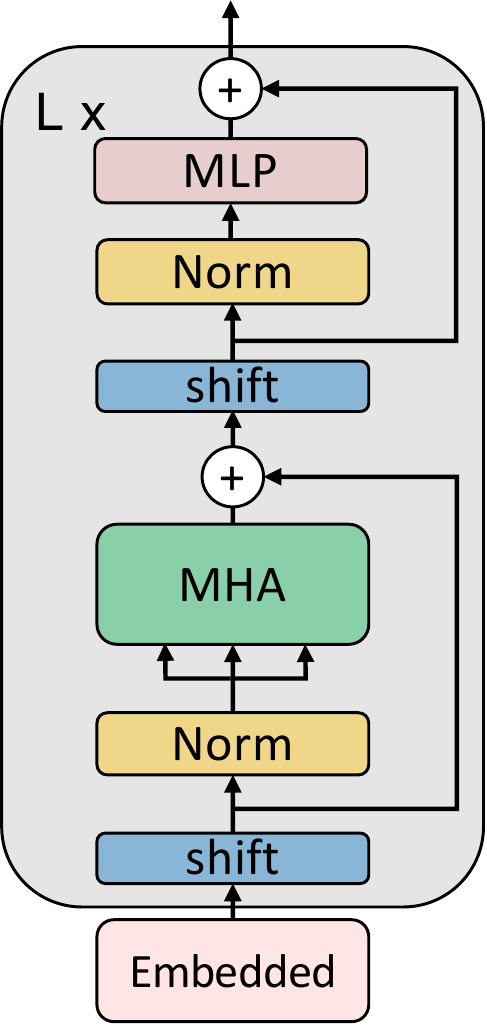}\\
    \subcaption{}
    \label{fig:tokenshift_block}
    \end{minipage}
    \hfil
    \begin{minipage}{.15\linewidth}
    \centering
    \includegraphics[width=\linewidth]{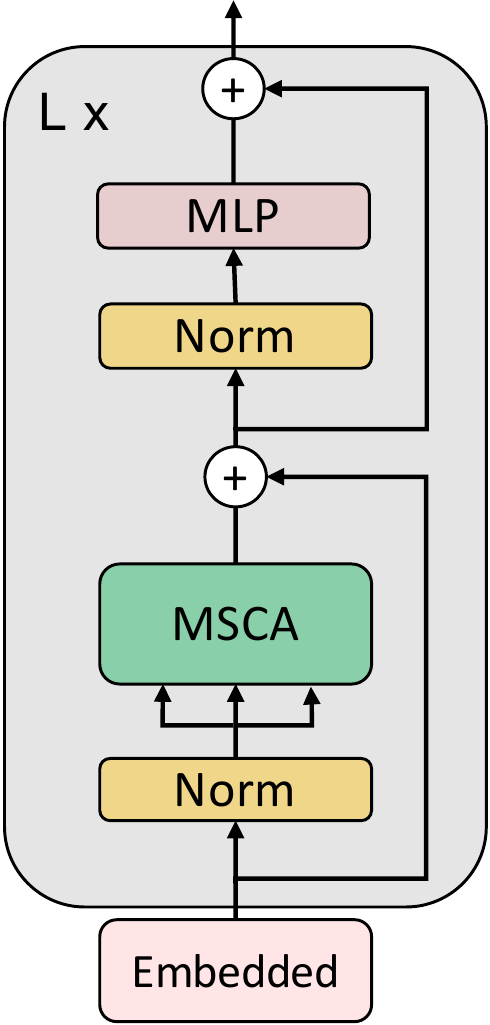}\\
    \subcaption{}
    \label{fig:MCSA_block}
    \end{minipage}

    \caption{
    Encoder blocks of
    (a) ViT,
    (b) TokenShift with two shift modules (in blue), and
    (c) the proposed method with MSCA (in green).
    }
    \label{Fig:overview}

\end{figure}

Figure \ref{fig:ViT_block} shows the original ViT encoder block
that has the Multi-head Self-Attention (MSA) module.
Fig.\ref{fig:tokenshift_block} shows the encoder block of TokenShift.
Two modules for shifting intermediate features are added to the original ViT block.
These modules shift a portion of the class tokens 
in the temporal direction by one time step forward and backward, similar to TSM.
Fig.\ref{fig:MCSA_block}
shows the encoder block of the proposed MSCA.
The difference is that MSA is replaced with the Multi-head Self/Cross-Attention (MSCA) module, and no additional modules exist.


\subsection{TokenShift and ViT}

In this section, we briefly review TokenShift \cite{Token-Shift-Transformer} which 
is based on ViT \cite{ViT}.

\subsubsection{Input Patch Embedding}

Let an input video be $x \in \R^{T \times 3 \times H \times W}$,
where $T$ is the number of frames in the video clip,
and $H, W$ are the height and width of the frame.
Each frame is divided into patches of
size $P \times P$ pixels
and transformed into a tensor $\hat{x} = [x_0^1, \ldots, x_0^N] \in \R^{T \times N \times d}$,
where $x_0^i \in \R^{T \times d}$ denotes the $i$-th patch,
$N = \frac{HW}{P^2}$ is the number of patches
of dimension $d = 3 P^2$.

The input patch $x_0^i$ is then transformed by the embedding matrix $E \in \R^{d \times D}$ and the positional encoding $E_\mathrm{pos}$ as follows:
\begin{align}
    z_0 &= [c_0, x_0^1 E, x_0^2 E, \ldots, x_0^N E] + E_\mathrm{pos},
\end{align}
where $c_0 \in \R^{T \times D}$ is the class token.
This patch embedding $z_0 \in \R^{T \times (N+1) \times D}$ is the input to the first encoder block.

\subsubsection{Encoder Block}

Let $z_\ell$ be the input to the $\ell$-th encoder block.
The output $z_\ell$
of the block can be expressed as follows:
\begin{align}
    z'_\ell &= \mathrm{MSA}(\mathrm{LN}(z_{\ell - 1})) + z_{\ell - 1}
    \\
    z_\ell &= \mathrm{MLP}(\mathrm{LN}(z'_\ell)) + z'_\ell,
\end{align}
where LN is the layer normalization,
MSA is the multi-head self-attention,
and MLP is the multi-layer perceptron.
In the following, $z_{\ell,t, n, d}$ denotes the element at $(t,n,d)$ in $z_\ell$.

\subsubsection{Shift Modules}

TokenShift inserts two shift modules in the block as follows;
\begin{align}
    z'_{\ell - 1} &= \mathrm{Shift}(z_{\ell - 1}) \\
    z''_\ell &= \mathrm{MSA}(\mathrm{LN}(z'_{\ell - 1})) + z'_{\ell - 1} \\
    z'''_\ell &= \mathrm{Shift}(z''_\ell) \\
    z_\ell &= \mathrm{MLP}(\mathrm{LN}(z'''_\ell)) + z'''_\ell.
\end{align}
The shift modules take
the input $z_\mathrm{in} \in R^{T \times (N+1) \times D}$
and compute the output $z_\mathrm{out}$ of the same size
by shifting the part of $z_\mathrm{in}$ corresponding to the class tokens
($z_{\mathrm{in},t,0,d}$,
the first elements of the second dimension of $z_\mathrm{in}$)
while leaving the other parts untouched.
This is implemented by the following assignments;
\begin{align}
    z_{\mathrm{out}, t, 0, d}
        &= 
    \begin{cases}
        z_{\mathrm{in}, t-1, 0, d}, & \text{$1 < t \le T, 1 \le d < D_b$}\\
        z_{\mathrm{in}, t+1, 0, d}, & \text{$1 \le t < T, D_b \le d < D_b+D_f$}\\
        z_{\mathrm{in}, t, 0, d},   & \text{$\forall t, D_b+D_f \le d \le D$}\\
    \end{cases}\\
    z_{\mathrm{out}, t, n, d}
        &= z_{\mathrm{in}, t, n, d}, \hspace{2.7em} \text{$\forall t, 1 \le n < N, \forall d$}
\end{align}
The first equation shifts the class tokens;
along the channel dimension $D$,
the backward shift to $t-1$ is done for the first $D_\mathrm{b}$ channels,
the forward shift to $t+1$ is done for the next  $D_\mathrm{f}$,
and no shift for the rest channels.
The second equation passes through features other than the class tokens.

\subsection{MSCA}

The proposed method replaces MSA with MSCA in the original block;
\begin{align}
    z'_\ell &= \mathrm{MSCA}(\mathrm{LN}(z_{\ell - 1})) + z_{\ell - 1}
    \\
    z_\ell &= \mathrm{MLP}(\mathrm{LN}(z'_\ell)) + z'_\ell.
\end{align}

In the following, we first describe MSA, and then define MSCA.

\subsubsection{MSA}

The original MSA computes the key $K^{(t)}$, query $Q^{(t)}$, and value $V^{(t)}$
for the portion $z^{(t)} \in \R^{(N+1) \times D}$ of the input feature $z \in \R^{T \times (N+1) \times D}$ at time $t$,
as follows;
\begin{align}
    K^{(t)}, Q^{(t)}, V^{(t)} &= z^{(t)} [W_k, W_q, W_v],
\end{align}
where $W_k, W_q, W_v \in \R^{D \times D}$ are embedding matrices,
and $z = [z^{(1)}, \ldots, z^{(T)}]$.
These are used to compute the $i$-th attention head;
\begin{align}
    \head_i^{(t)}
    &=
    a( Q^{(t)}_i, K^{(t)}_i)
    V^{(t)}_i
    \in \R^{(N+1) \times D/h},
\end{align}
at time $t$ for $i=1,\ldots,h$,
where the attention $a$ is
\begin{align}
    a(Q, K)
    &=
    \softmax (Q K^T / \sqrt{D}),
\end{align}
and $Q_i^{(t)} \in \R^{(N+1) \times D/h}$ is the part of $Q^{(t)}$ corresponding to $i$-th head
\begin{align}
Q^{(t)} = [Q_1^{(t)}, \ldots, Q_i^{(t)}, \ldots, Q_h^{(t)}],
\end{align}
and $K_i^{(t)}, V_i^{(t)}$ are the same.
These heads are finally stacked to form
\begin{align}
    \mathrm{MSA}(z^{(t)})
    &=
    [\head_1^{(t)}, \ldots, \head_{h}^{(t)} ].
\end{align}

In MSA, patches in $t$-th frame are attended from other patches of the same frame at time $t$,
which means that there are no temporal interactions between frames.

\begin{figure}[t]

    \centering
    
    \includegraphics[width=.6\linewidth]{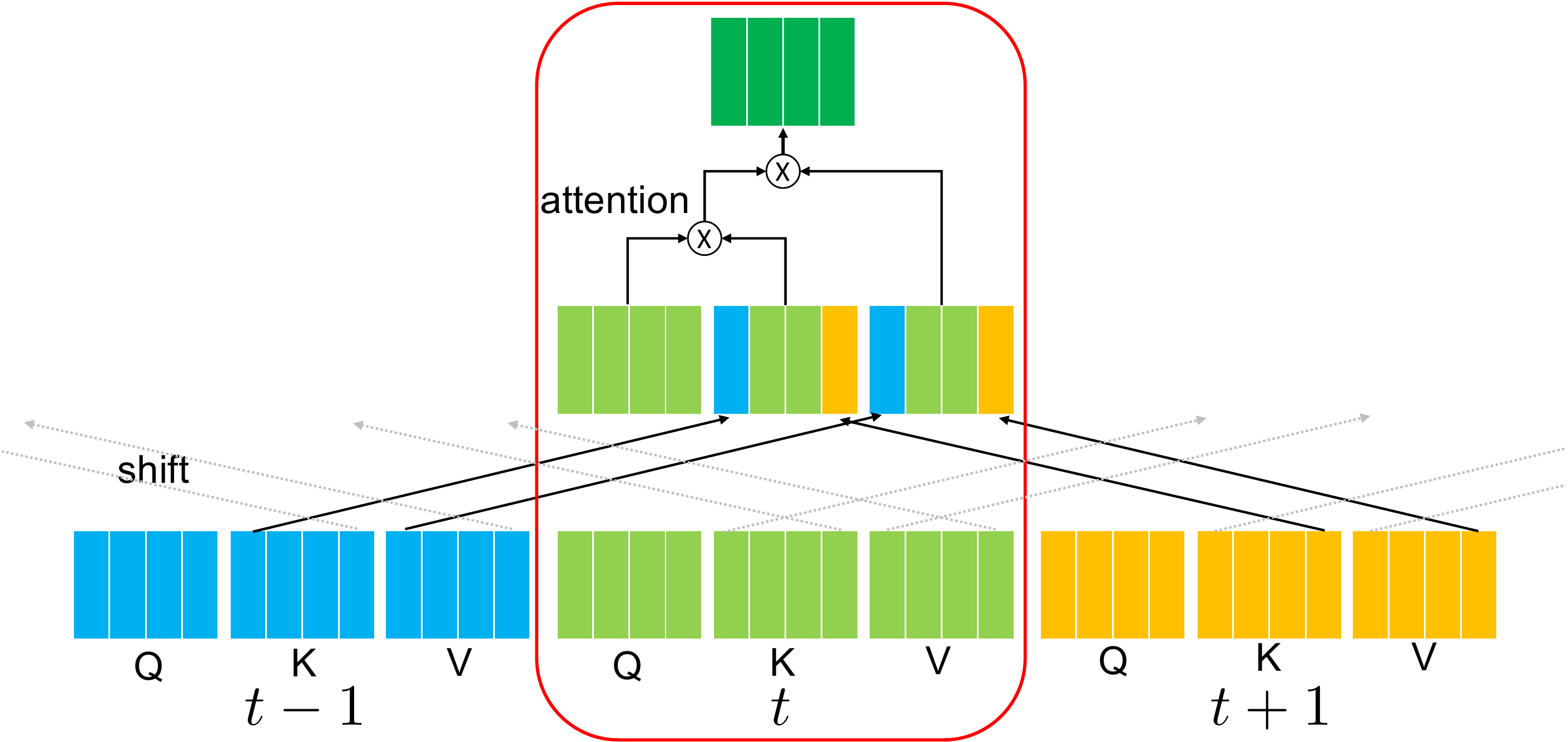}

    \caption{
    Shit operation at $T = t$ in the MSCA-KV model
    }
    \label{Fig:MSCA-KV}

\end{figure}

\begin{figure}[t]

    \centering
    


    \begin{minipage}{.3\linewidth}
    \centering
    \includegraphics[width=\linewidth]{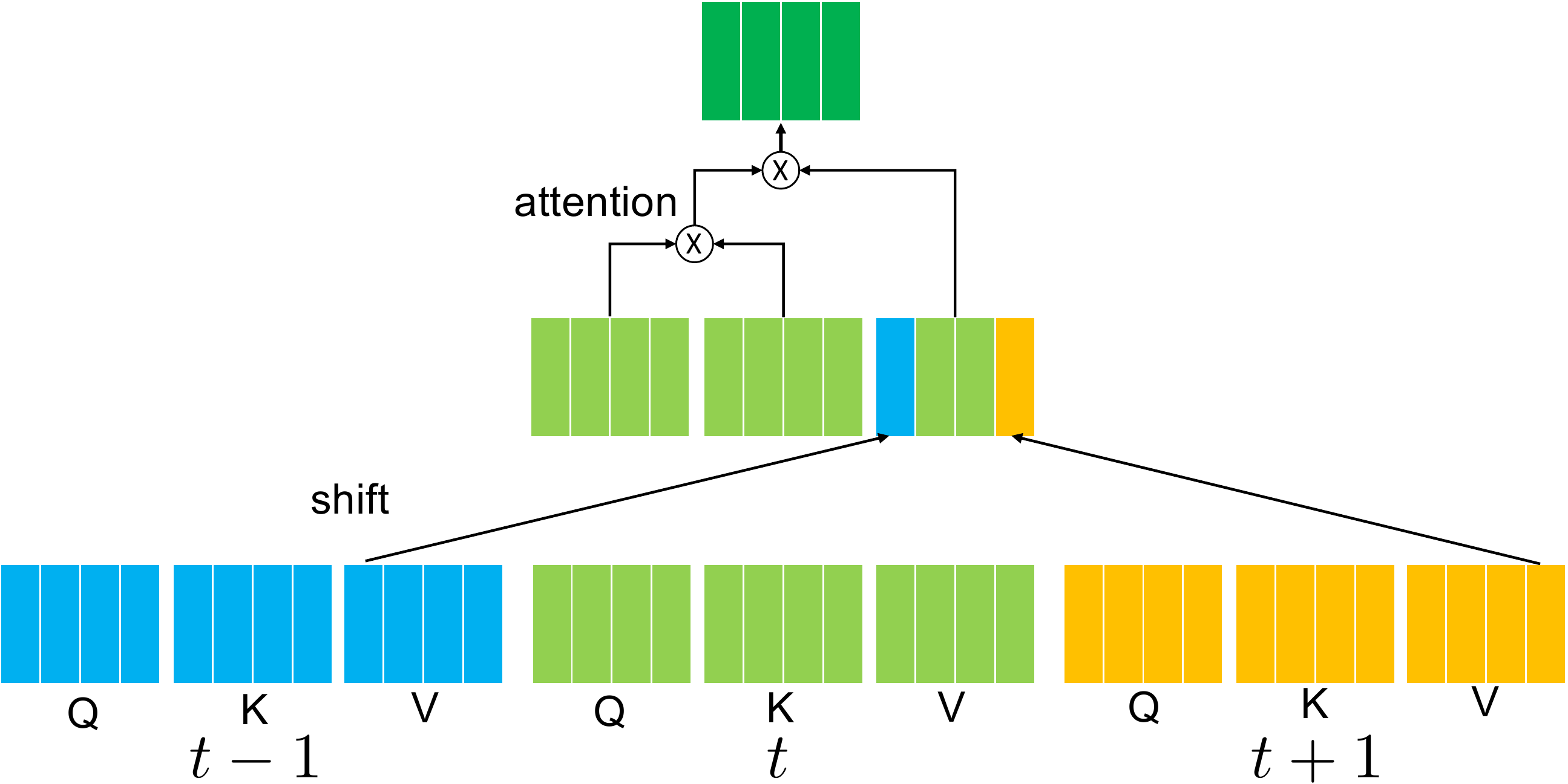}\\
    \subcaption{}
    \label{Fig:MSCA-V}
    \end{minipage}
    \hfill            
    \begin{minipage}{.3\linewidth}
    \centering
    \includegraphics[width=\linewidth]{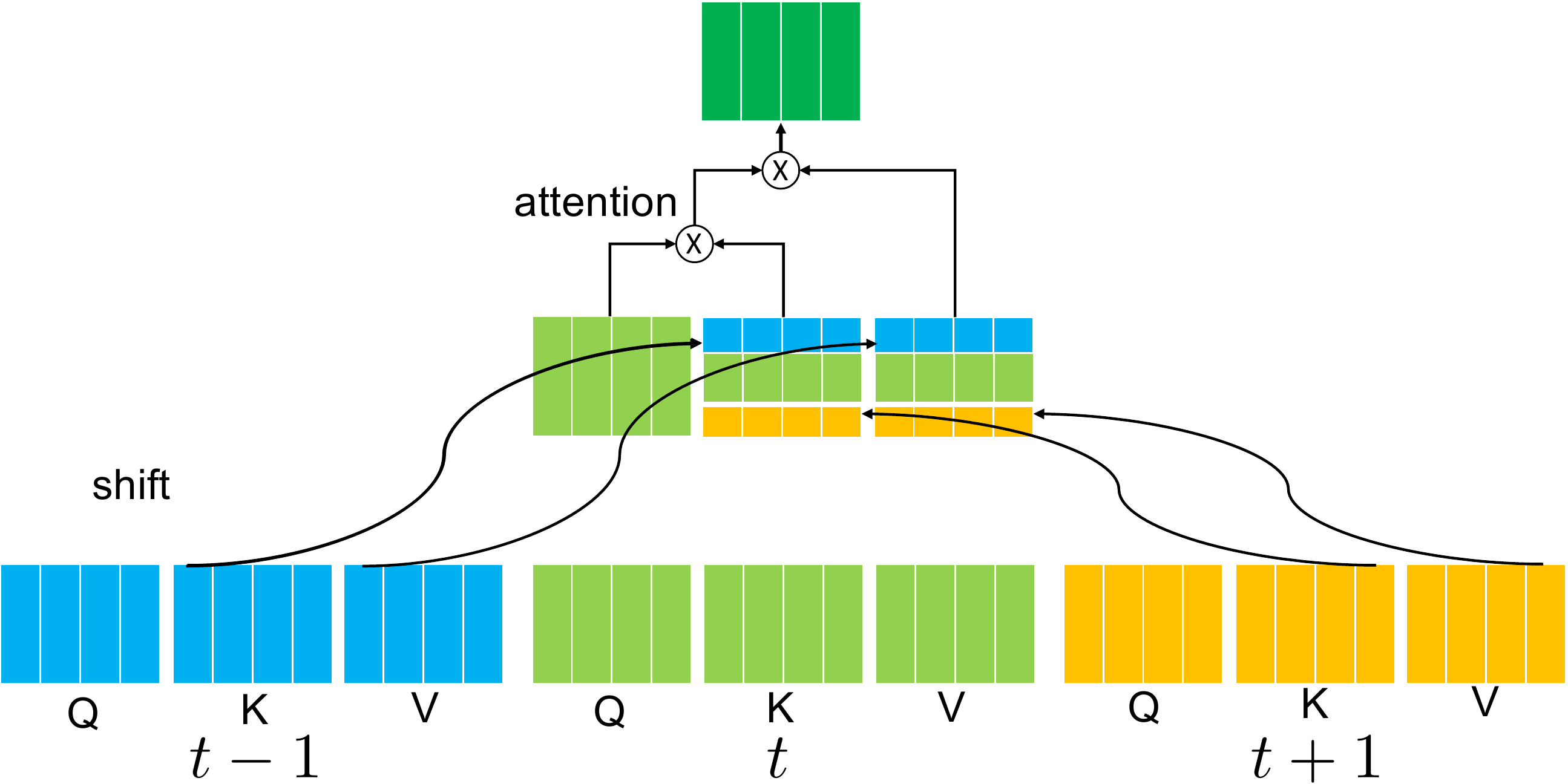}\\
    \subcaption{}
    \label{Fig:MSCA-pKV}
    \end{minipage}
    \hfill
    \begin{minipage}{.3\linewidth}
    \centering
    \includegraphics[width=\linewidth]{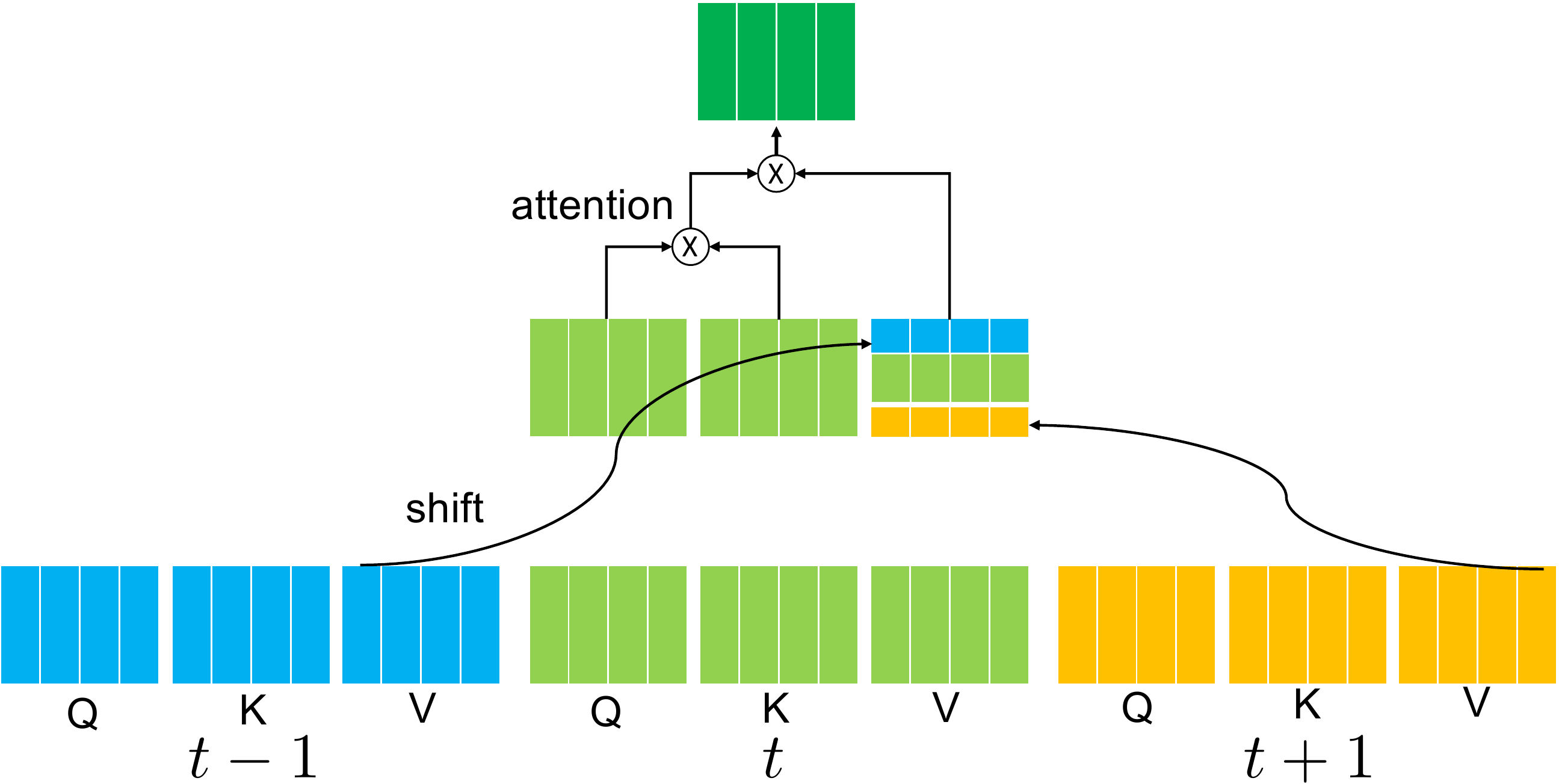}\\
    \subcaption{}
    \label{Fig:MSCA-pV}
    \end{minipage}

    \caption{Shit operations of
    (a) MSCA-V,
    (b) MSCA-pKV, and
    (c) MSCA-pV.
    }
    \label{Fig:MSCA}

\end{figure}

\subsubsection{MSCA-KV}

The proposed MSCA computes the attention across frames,
that is,
patches in $t$-th frame are also attended
from patches in frames at time $t+1$ and $t-1$. 
This can be done with shift operations;
after generating $Q, K, V$ at each frame,
these are exchanged with the neighbor frames.

There are choices of which of $Q, K$, or $V$ to be shifted.
A possible choice is to shift $K$ and $V$,
and the query in the current frame is attended by key-value pairs in other frames.
This is expressed as follows;
\begin{align}
    \head_i^{(t)}
    &=
    \begin{cases}
        a(Q^{(t)}_i, K^{(t-1)}_i) V^{(t-1)}_i    & 1   \le i < h_b\\
        a(Q^{(t)}_i, K^{(t+1)}_i) V^{(t+1)}_i    & h_b \le i < h_b+h_f \\
        a(Q^{(t)}_i, K^{(t)}_i)   V^{(t)}_i      & h_b+h_f \le i \le h.
    \end{cases}
\end{align}
Here, we have 
the first $h_b$ heads with the backward shift,
the next $h_f$ heads with the forward shift,
and the rest heads with no shift.
We call this MSCA-KV, and Figure \ref{Fig:MSCA-KV} shows the diagram of the shift operation.
First, queries, keys and values are computed at each frame,
and then some of them are shifted before computing the attention.
The solid arrows indicate 
the key-value shift from time $t-1$ and $t+1$.
As in the same way,
the key-value shift from the current frame $t$ to $t-1$ and $t+1$
are shown in dotted arrows.
There are shifts in all other frames at the same time in the same manner.

\subsubsection{MSCA-V}

Another choice of shift is shown 
in Fig. \ref{Fig:MSCA-V}. Here,
$Q$ and $K$ are not shifted, but only $V$ is shifted as follows;
\begin{align}
    \head_i^{(t)}
    &=
\begin{cases}
    a(Q^{(t)}_i, K^{(t)}_i) V^{(t-1)}_i
    & 1 \le i < h_b \\
    a(Q^{(t)}_i, K^{(t)}_i) V^{(t+1)}_i
    & h_b \le i < h_b+h_f\\
    a(Q^{(t)}_i, K^{(t)}_i) V^{(t)}_i
    & h_b+h_f \le i \le h.
\end{cases}
\end{align}
This might not be common because the key and value are now separated
and taken from different frames.
However, this makes sense for modeling temporal interactions because
the values (which are mixed by the attention weights)
come from different frames while the attention is computed in the current frame.
We call this version MSCA-V.
Therefore, in addition to shifting $V$ in this way,
there are seven possible combinations; Q, K, V, QK, KV, QV, and QKV.
We compared these variants in the experiments.
Note that MSCA-QKV is equivalent to a simple feature shifting
because attended features are computed in each frame and then shifted.

\subsubsection{MSCA-pKV}

All of the above seven variants perform shift operations along the head (or channel) dimension $D$,
but similar variants of shift are also possible for the patch dimension $N+1$.

The shapes of $K^{(t)}, Q^{(t)}, V^{(t)}$ are $\R^{(N+1) \times D}$,
and the first dimension is for patches
while the second is for heads.
As in the same way the the shift along the head dimension,
we have a variation of shift operations along the patch
dimension as shown in Fig. \ref{Fig:MSCA-pKV}.

First, $K$ and $V$ are expressed as stacks of keys and values of patches at different frames as follows;
\begin{align}
    K^{(t)} &= [K^{(t)}_0, K^{(t)}_1, \ldots, K^{(t)}_N] \\
    V^{(t)} &= [V^{(t)}_0, V^{(t)}_1, \ldots, V^{(t)}_N],
\end{align}
where $K^{(t)}_n, V^{(t)}_n \in \R^D$ are the key and value of patch $n$ at time $t$.

Then, keys of some patches in the current frame are shifted to form $K'$;
\begin{align}
    K'^{(t)}_n
    &=
\begin{cases}
    K^{(t-1)}_n & 0 \le n < N_b \\
    K^{(t+1)}_n & N_b \le n < N_b+N_f\\
    K^{(t)}_n   & N_b+N_f \le n \le N,
\end{cases}
\end{align}
and also $V'$ in the same way.
Finally, the $i$-th head is computed as follows
\begin{align}
    \head_i^{(t)} &= 
        a(Q^{(t)}_i, K'^{(t)}_i)
        V'^{(t)}_i.
\end{align}
We refer to this version as MSCA-pKV.

\subsubsection{MSCA-pV}

Like as MSCA-V,
a variant of the shift in the patch direction with $V$ only
can be also considered as shown in Fig. \ref{Fig:MSCA-pV}, by the following shift;
\begin{align}
    \head_i^{(t)} &=
    a(Q^{(t)}_i, K^{(t)}_i)
    V'^{(t)}_i.
\end{align}
As before, there are seven variants,
and we call these MSCA-pV, and so on.

\section{Experimental results}

\subsection{Setup}

Kinetics400 \cite{Kinetics}
was used to train and evaluate the proposed method. 
This dataset consists of a training set of 22k videos, a validation set of 18k videos, with 400 categories of human actions.
Each video was collected from Youtube, and the portion corresponding to each category was cropped to a length of 10 seconds.


We used a 2D ViT pre-trained on ImageNet21k \cite{ImageNet}
with $h=12$ heads, 12 encoder blocks, and patches of size $P=16$
and $D = 768 = 3 \times 16 \times 16$
(these parameters are the same for TokenShift and MSCA models).
For action recognition,
ViT was applied to each frame and resulting frame-wise features were aggregated by temporal averaging
(this is referred to as ViT in the experiment and in \cite{Token-Shift-Transformer}).
We compared this ViT, TokenShift
and the proposed method.
Note that we report the performance of TokenShift based on our reproduction
using the author's code.%
\def\UrlFont{\ttfamily\scriptsize}%
\footnote{\url{https://github.com/VideoNetworks/TokShift-Transformer}}

For training, we used
the same settings as in \cite{Token-Shift-Transformer}.
Input clips were of 8 frames with stride of 32 frames (starting frames were randomly chosen).
Frames were flipped horizontally at a probability of 50\%,
and the short side was randomly resized in the range of [244, 330] pixels while maintaining the aspect ratio, 
then a random $224 \times 224$ pixel rectangle was cropped (therefore the number of patches is $N = 196 = 14 \times 14$).
In addition, brightness change, saturation change, gamma correction, and hue correction were applied to frames, each at the probability of 10\%.
The number of epochs was set to 12,
the optimizer to SDG with momentum of 0.9 and no weight decay.
The initial learning rate was set to 0.1,
and decayed by a factor of 10 at 10th epoch.
The batch size was set to 42,
and 21 batches were trained on each of two GPUs.
Gradient updates were performed
once every 10 iterations, so the effective batch size was 420.


We used the multi-view test \cite{Nonlocal}.
From a validation video, one clip was sampled as in training,
and this was repeated 10 times to sample 10 clips. 
Each clip was resized to 224 pixels on its short side while maintaining the aspect ratio,
and cropped to $224\times224$ at the right, center, and left. 
The results of these 30 clips (views) were averaged to compute a single prediction score.

\subsection{The amount of shift of MSCA-KV}

\begin{table}[t]

  \centering
  \caption{
    The performance of MSCA-KV for the validation sets of Kinetics400.
    Note that the zero shift means a naive frame-wise ViT.
    The column ``shift'' means the percentage of the shifted dimensions to the total dimensions $D$.
  }
  \label{Tab:KV_shift_acc}
  
  \medskip
  
\begin{tabular}{cccc|cc}
model & heads & $h_b,h_f$ & shift \% & top-1          & top-5          \\ \hline
ViT & 0     & 0         & 0 (ViT)  & 75.65          & 92.19          \\
MSCA-KV & 2 & 1         & 16.7     & \textbf{76.47} & \textbf{92.88} \\
& 4     & 2         & 33.3     & 76.07          & 92.61          \\
& 6     & 3         & 50.0     & 75.66          & 92.30          \\
& 8     & 4         & 66.7     & 74.72          & 91.91         
\end{tabular}

\end{table}

We first investigate the effect of the number of heads to be shifted.
Table \ref{Tab:KV_shift_acc} shows the performance of MSCA-KV.
The best performance was obtained when only two heads (each for forward and backward)
which corresponds to shifting $2/12 = 16.7\%$ of the channels.
As the number of shifted heads increased, the performance decreased,
suggesting that shifting a few heads is sufficient while most heads need not to be shifted.
This observation coincides with the conclusions of TokenShift \cite{Token-Shift-Transformer},
which shows that the shift of the class token only is enough,
and also TSM \cite{Lin_2019_ICCV}, which used the shift of $1/4 = 25\%$ of the channels (each 1/8 for forward and backward).

In the following experiments, we used shifting two heads for all MSCA variations.

\subsection{Comparison of MSCA variations shifting in the head direction}

\begin{table}[t]
  \centering

\begin{minipage}[t]{.4\linewidth}
  \centering
  \caption{
    The performance of MSCA variants shifting in the head direction.
  }
  \label{Tab:head_shift}
  \medskip
  \begin{tabular}{c|cc}  
    model & top-1 & top-5 \\ \hline
    TokenShift & 76.37 & 92.82 \\
    ViT & 75.65 & 92.19 \\ \hline
    MSCA-Q & 75.84 & 92.57 \\
    MSCA-K & 75.76 & 92.13 \\
    MSCA-V & 75.58 & 92.39 \\
    MSCA-QK& 75.47 & 92.32 \\
    MSCA-KV & \textbf{76.47} & \textbf{92.88} \\
    MSCA-QV& 75.57 & 92.43 \\
    MSCA-QKV& 75.78 & 92.37 \\
  \end{tabular}
\end{minipage}
\hfil
\begin{minipage}[t]{.53\linewidth}
  \centering
  \caption{
    The performance with different numbers of blocks with MSCA modules of KV shift.
    }
  \label{Tab:KV_shift_encoder}
  \medskip
  \begin{tabular}{@{}c@{}c@{}c|cc}  
    model & \# MSCA  & \# MSA & top-1 & top-5 \\ \hline
    ViT & 0         & 12     & 75.65 & 92.19 \\
        & 4        & 8      & 75.67 & 92.19 \\ 
        & 8        & 4      & 76.40 & 92.77 \\
  MSCA-KV & 12 & 0      & \textbf{76.47} & \textbf{92.88} \\
  \end{tabular}
\end{minipage}

\end{table}

Table \ref{Tab:head_shift} shows the performance of the MSCA model variants with shift operations in the head direction.
Among them, MSCA-KV performed the best, outperforming others by at least 0.5\%.
Other variants performed as same as the baseline ViT,
indicating that the shift operation is not working effectively in such variations.

\subsection{The number of encoder blocks with MSCA}

The proposed method replaces MSA modules in 12 encoder blocks in ViT with MSCA modules.
However, it is not obvious that modules of all blocks need to be replaced.
Table \ref{Tab:KV_shift_encoder} shows the results
when we replaced MSA modules  with MSCA in some blocks near the end (or top) of the network.
All 12 replacements correspond to MSCA-KV, and 0 (no MCSA) is ViT.
Using four MSCA modules showed no improvement,
while using 8 MSCA modules performed almost the same as MSCA-KV with all 12 MSCA modules.
This is because the input video clip consists of 8 frames,
and shifting more than 8 times with MSCA modules ensures that the temporal information from all frames is available for the entire network.
Therefore, it would be necessary to use at least as many MSCA modules as the number of frames in the input clip.

\subsection{Comparison of MSCA variations shifting in the patch direction}


\begin{table}[t]
  \centering

\begin{minipage}[t]{.45\linewidth}
\centering
  \caption{
    The performance of MSCA-pKV.    
    The column ``shift'' means the percentage of the shifted patches to the total patches $N + 1$.
  }
  \label{Tab:pKV_shift_acc}
\medskip
\begin{tabular}{ccc|cc}
patches & $N_b, N_b$ & shift \%   & top-1          & top-5          \\ \hline
0       & 0          & 0 (ViT) & 75.65          & 92.19          \\
8       & 4          & 4.1     & 76.28          & 92.49          \\
16      & 8          & 8.1     & \textbf{76.35} & \textbf{92.91} \\
32      & 16         & 16.2    & 76.07          & 92.77          \\
48      & 24         & 24.4    & 75.84          & 92.49          \\
64      & 32         & 32.5    & 75.29          & 92.04         
\end{tabular}
\end{minipage}
\hfil
\begin{minipage}[t]{.45\linewidth}
\centering
  \caption{
    The performance of MSCA models shifting in the patch direction.
  }
  \label{Tab:patch_shift}
  \medskip
  \begin{tabular}{c|cc}  
    model & top-1 & top-5 \\ \hline
    TokenShift & \textbf{76.37} & 92.82 \\
    ViT & 75.65 & 92.19 \\ \hline
    MSCA-pQ & 75.75 & 92.10 \\
    MSCA-pK & 75.69 & 92.24 \\
    MSCA-pV & 75.84 & 92.43 \\
    MSCA-pQK& 75.50 & 91.99 \\
    MSCA-pKV & 76.35 & \textbf{92.91} \\
    MSCA-pQV& 75.83 & 92.35 \\
    MSCA-pQKV& 75.73 & 92.21 \\
  \end{tabular}
\end{minipage}

\end{table}

Table \ref{Tab:pKV_shift_acc} shows the performance of MSCA-pKV with different amount of shift.
The results indicate that, again, a small amount of shift is enough for a better performance
and the best performance was obtained when 16 patches were shifted.
The performance decreases for a smaller amount of shift,
approaching the performance of ViT with no shift.
In the experiments below, we used the shift of 16 patches.

Table \ref{Tab:patch_shift} shows the performance of MSCA variants with shifting in the patch direction.
Just like as MSCA-KV was the best for shifting in the head direction,
MSCA-pKV has the best performance here, while it is just comparable to TokenShift.
An obvious drawback of this approach is the first layer;
the patches to be shifted were fixed to the first $N_b$ and $N_f$ patches
in the order of the patch index, which is irrelevant to the content of the frame.
This might be mitigated by not using MSCA modules in the first several layers.

\section{Conclusions}

In this paper, we proposed MSCA, a ViT-based action recognition model
that replaces MSA modules in the encoder blocks.
The MSCA modules compute the attention by shifting of the key, query, and value
for temporal interaction between frames.
Experimental results using Kinetics400 showed that the proposed method
is effective for modeling spatio-temporal features and performs
better than a naive ViT and TokenShift.
Future work includes evaluations on other datasets and comparisons
with similar methods such as Space-time Mixing Attention \cite{bulat2021spacetime}.

%% file: accv2022submission.bbl
\begin{thebibliography}{10}

\bibitem{ViT}
Dosovitskiy, A., Beyer, L., Kolesnikov, A., Weissenborn, D., Zhai, X.,
  Unterthiner, T., Dehghani, M., Minderer, M., Heigold, G., Gelly, S.,
  Uszkoreit, J., Houlsby, N.:
\newblock An image is worth 16x16 words: Transformers for image recognition at
  scale.
\newblock In: International Conference on Learning Representations. (2021)

\bibitem{CLIP}
Radford, A., Kim, J.W., Hallacy, C., Ramesh, A., Goh, G., Agarwal, S., Sastry,
  G., Askell, A., Mishkin, P., Clark, J., Krueger, G., Sutskever, I.:
\newblock Learning transferable visual models from natural language
  supervision.
\newblock CoRR \textbf{abs/2103.00020} (2021)

\bibitem{DALL-E}
Ramesh, A., Pavlov, M., Goh, G., Gray, S., Voss, C., Radford, A., Chen, M.,
  Sutskever, I.:
\newblock Zero-shot text-to-image generation.
\newblock In Meila, M., Zhang, T., eds.: Proceedings of the 38th International
  Conference on Machine Learning. Volume 139 of Proceedings of Machine Learning
  Research., PMLR (2021)  8821--8831

\bibitem{ViViT}
Arnab, A., Dehghani, M., Heigold, G., Sun, C., Lu\v{c}i\'c, M., Schmid, C.:
\newblock Vivit: A video vision transformer.
\newblock In: Proceedings of the IEEE/CVF International Conference on Computer
  Vision (ICCV). (2021)  6836--6846

\bibitem{VidTr}
Li, X., Zhang, Y., Liu, C., Shuai, B., Zhu, Y., Brattoli, B., Chen, H., Marsic,
  I., Tighe, J.:
\newblock Vidtr: Video transformer without convolutions.
\newblock CoRR \textbf{abs/2104.11746} (2021)

\bibitem{Video-Transformer-Network}
Girdhar, R., Carreira, J., Doersch, C., Zisserman, A.:
\newblock Video action transformer network.
\newblock In: Proceedings of the IEEE/CVF Conference on Computer Vision and
  Pattern Recognition (CVPR). (2019)

\bibitem{TimeSformer}
Bertasius, G., Wang, H., Torresani, L.:
\newblock Is space-time attention all you need for video understanding?
\newblock In: Proceedings of the International Conference on Machine Learning
  (ICML). (2021)

\bibitem{STAM}
Sharir, G., Noy, A., Zelnik{-}Manor, L.:
\newblock An image is worth 16x16 words, what is a video worth?
\newblock CoRR \textbf{abs/2103.13915} (2021)

\bibitem{Karpathy_2014_CVPR}
Karpathy, A., Toderici, G., Shetty, S., Leung, T., Sukthankar, R., Fei-Fei, L.:
\newblock Large-scale video classification with convolutional neural networks.
\newblock In: Proceedings of the IEEE Conference on Computer Vision and Pattern
  Recognition (CVPR). (2014)

\bibitem{Donahue_2015_CVPR}
Donahue, J., Anne~Hendricks, L., Guadarrama, S., Rohrbach, M., Venugopalan, S.,
  Saenko, K., Darrell, T.:
\newblock Long-term recurrent convolutional networks for visual recognition and
  description.
\newblock In: Proceedings of the IEEE Conference on Computer Vision and Pattern
  Recognition (CVPR). (2015)

\bibitem{C3D}
Tran, D., Bourdev, L., Fergus, R., Torresani, L., Paluri, M.:
\newblock Learning spatiotemporal features with 3d convolutional networks.
\newblock In: Proceedings of the IEEE International Conference on Computer
  Vision (ICCV). (2015)

\bibitem{I3D}
Carreira, J., Zisserman, A.:
\newblock Quo vadis, action recognition? a new model and the kinetics dataset.
\newblock In: Proceedings of the IEEE Conference on Computer Vision and Pattern
  Recognition (CVPR). (2017)

\bibitem{R3D}
Hara, K., Kataoka, H., Satoh, Y.:
\newblock Can spatiotemporal 3d cnns retrace the history of 2d cnns and
  imagenet?
\newblock In: Proceedings of the IEEE Conference on Computer Vision and Pattern
  Recognition (CVPR). (2018)  6546--6555

\bibitem{Kinetics}
Kay, W., Carreira, J., Simonyan, K., Zhang, B., Hillier, C., Vijayanarasimhan,
  S., Viola, F., Green, T., Back, T., Natsev, P., Suleyman, M., Zisserman, A.:
\newblock The kinetics human action video dataset.
\newblock CoRR \textbf{abs/1705.06950} (2017)

\bibitem{UCF101}
Soomro, K., Zamir, A.R., Shah, M.:
\newblock Ucf101: A dataset of 101 human actions classes from videos in the
  wild.
\newblock CoRR \textbf{abs/1212.0402} (2012)

\bibitem{Lin_2019_ICCV}
Lin, J., Gan, C., Han, S.:
\newblock Tsm: Temporal shift module for efficient video understanding.
\newblock In: Proceedings of the IEEE/CVF International Conference on Computer
  Vision (ICCV). (2019)

\bibitem{CNN-Transformer-CVPR2019}
Chen, W., Xie, D., Zhang, Y., Pu, S.:
\newblock All you need is a few shifts: Designing efficient convolutional
  neural networks for image classification.
\newblock In: Proceedings of the IEEE/CVF Conference on Computer Vision and
  Pattern Recognition (CVPR). (2019)

\bibitem{Wu_2018_CVPR}
Wu, B., Wan, A., Yue, X., Jin, P., Zhao, S., Golmant, N., Gholaminejad, A.,
  Gonzalez, J., Keutzer, K.:
\newblock Shift: A zero flop, zero parameter alternative to spatial
  convolutions.
\newblock In: Proceedings of the IEEE Conference on Computer Vision and Pattern
  Recognition (CVPR). (2018)

\bibitem{ResNet2015}
He, K., Zhang, X., Ren, S., Sun, J.:
\newblock Deep residual learning for image recognition.
\newblock In: Proceedings of the IEEE Conference on Computer Vision and Pattern
  Recognition (CVPR). (2016)

\bibitem{Gated-Shift-Network}
Sudhakaran, S., Escalera, S., Lanz, O.:
\newblock {Gate-Shift Networks for Video Action Recognition}.
\newblock In: IEEE/CVF Conference on Computer Vision and Pattern Recognition
  (CVPR). (2020)

\bibitem{LGTSM}
Chang, Y.L., Liu, Z.Y., Lee, K.Y., Hsu, W.:
\newblock Learnable gated temporal shift module for deep video inpainting".
\newblock BMVC (2019)

\bibitem{RubiksNet}
Fan, L., Buch, S., Wang, G., Cao, R., Zhu, Y., Niebles, J.C., Fei-Fei, L.:
\newblock Rubiksnet: Learnable 3d-shift for efficient video action recognition.
\newblock In Vedaldi, A., Bischof, H., Brox, T., Frahm, J.M., eds.: Computer
  Vision -- ECCV 2020, Cham, Springer International Publishing (2020)

\bibitem{Token-Shift-Transformer}
Zhang, H., Hao, Y., Ngo, C.W.
\newblock In: Token Shift Transformer for Video Classification. Association for
  Computing Machinery, New York, NY, USA (2021)  917–925

\bibitem{two-stream}
Simonyan, K., Zisserman, A.:
\newblock Two-stream convolutional networks for action recognition in videos.
\newblock In Ghahramani, Z., Welling, M., Cortes, C., Lawrence, N., Weinberger,
  K.Q., eds.: Advances in Neural Information Processing Systems. Volume~27.,
  Curran Associates, Inc. (2014)

\bibitem{P3D}
Qiu, Z., Yao, T., Mei, T.:
\newblock Learning spatio-temporal representation with pseudo-3d residual
  networks.
\newblock In: Proceedings of the IEEE International Conference on Computer
  Vision (ICCV). (2017)

\bibitem{S3D}
Zhang, D., Dai, X., Wang, X., Wang, Y.F.:
\newblock S3d: Single shot multi-span detector via fully 3d convolutional
  network.
\newblock In: Proceedings of the British Machine Vision Conference (BMVC).
  (2018)

\bibitem{R(2+1)D}
Tran, D., Wang, H., Torresani, L., Ray, J., LeCun, Y., Paluri, M.:
\newblock A closer look at spatiotemporal convolutions for action recognition.
\newblock In: Proceedings of the IEEE Conference on Computer Vision and Pattern
  Recognition (CVPR). (2018)

\bibitem{tokenlearner}
Ryoo, M.S., Piergiovanni, A., Arnab, A., Dehghani, M., Angelova, A.:
\newblock Tokenlearner: Adaptive space-time tokenization for videos.
\newblock In: Advances in Neural Information Processing Systems (NeurIPS).
  (2021)

\bibitem{bulat2021spacetime}
Bulat, A., Perez-Rua, J.M., Sudhakaran, S., Martinez, B., Tzimiropoulos, G.:
\newblock Space-time mixing attention for video transformer.
\newblock In Beygelzimer, A., Dauphin, Y., Liang, P., Vaughan, J.W., eds.:
  Advances in Neural Information Processing Systems. (2021)

\bibitem{ImageNet}
Deng, J., Dong, W., Socher, R., Li, L.J., Li, K., Fei-Fei, L.:
\newblock Imagenet: A large-scale hierarchical image database.
\newblock In: 2009 IEEE Conference on Computer Vision and Pattern Recognition.
  (2009)  248--255

\bibitem{Nonlocal}
Wang, X., Girshick, R., Gupta, A., He, K.:
\newblock Non-local neural networks.
\newblock In: Proceedings of the IEEE Conference on Computer Vision and Pattern
  Recognition (CVPR). (2018)

\end{thebibliography}
